\documentclass[12pt,a4paper]{article}
\usepackage[utf8]{inputenc}
\usepackage[T1]{fontenc}
\usepackage{geometry}
\usepackage{amsmath,amsfonts,amssymb}
\usepackage{graphicx}
\usepackage{booktabs}
\usepackage{longtable}
\usepackage{adjustbox}
\usepackage{siunitx}
\usepackage{microtype}
\usepackage{hyperref}
\usepackage{setspace}
\usepackage{parskip}
\usepackage{tabularx}
\usepackage{booktabs}
\usepackage{makecell}
\usepackage{adjustbox}
\usepackage{ragged2e}
\usepackage{caption}
\usepackage{float}
\usepackage[numbers]{natbib}
\usepackage{hyperref}
\usepackage[T1]{fontenc}
\usepackage[utf8]{inputenc}
\usepackage{lmodern}

\usepackage{geometry}
\usepackage{microtype}
\usepackage{amsmath,amsfonts,amssymb}
\usepackage{graphicx}
\usepackage{booktabs}
\usepackage{hyperref}
\usepackage[numbers]{natbib}
\usepackage{parskip}
\usepackage{titling}

\pretitle{%
\begin{center}
\rule{\linewidth}{2pt}\\[0.5em] 
\LARGE\bfseries  
}
\posttitle{%
\\[0.5em]\rule{\linewidth}{2pt} 
\end{center}\vspace{1em}
}

\preauthor{\begin{center}\large}
\postauthor{\par\end{center}}

\title{Evaluating Embedding Generalization:\\How LLMs, LoRA, and SLERP Shape Representational Geometry}

\author{
Siyaxolis Kabane\\[0.3em]
Department of Computer Science\\
University of Fort Hare\\[0.3em]
\texttt{202378066@ufh.ac.za}
}

\date{}

\begin{document}

\maketitle
\begin{abstract}
We investigate the generalization properties of dense text embeddings when the embedding backbone is a large language model (LLM) versus when it is a non-LLM encoder, and we study the extent to which spherical linear interpolation (SLERP) model-merging mitigates over-specialization introduced by task-specific adaptation (e.g., LoRA). To make the comparison concrete and domain-agnostic, we design a controlled suite of experiments in which models embed short numerical sequences and are evaluated on their ability to cluster and classify those sequences according to well-defined number-theoretic properties. Our experimental protocol compares four families of models: (1) non-LLM encoders trained from scratch or fine-tuned for embeddings, (2) LLM-based encoders adapted with parameter-efficient methods (LoRA), (3) LLM-based encoders with LoRA followed by model souping merging into the base weights, and (4) the same LoRA-adapted LLMs merged using SLERP across checkpoints or stages. We evaluate representational quality with clustering indices (Silhouette and Davies–Bouldin). We additionally analyze the use of kmeans labels to see if the embeddings encode any other information besides the one we are testing for. Empirically, we find that LLM-based backbones produce embeddings that better capture higher-order, compositional numeric patterns, but are prone to adapter dominance that degrades balanced generalization; SLERP merging consistently recovers base-model structure while retaining most task gains, yielding superior tradeoffs in clustering separability, and robustness compared to model souping or models that were not merged.

\end{abstract}

\section{Introduction}
\label{sec:intro}

Dense vector representations (``embeddings'') lie at the heart of modern natural language processing and information retrieval: they provide compact, continuous encodings that make semantic comparison, clustering, and retrieval efficient and effective. Early work demonstrated that unsupervised co-occurrence objectives produce remarkably useful fixed word vectors (e.g., GloVe), and that simple aggregation schemes can already yield strong sentence- and document-level representations under many conditions. \citep{pennington2014glove,arora2017simple}

More recent work moved beyond token-level vectors to learned sentence and text embeddings that are explicitly trained for semantic similarity and retrieval. Two influential directions are (i) adapting large pretrained transformer encoders into fast similarity encoders (e.g., Sentence-BERT) and (ii) contrastive/self-supervised objectives that substantially improve the isotropy and transfer of sentence representations (e.g., SimCSE). These methods made it practical to compute high-quality sentence embeddings at scale and established benchmarks and recipes that are now widely used across tasks. \citep{reimers2019sentence,gao2021simcse}

Benchmarking and careful evaluation have been critical to measuring progress and to revealing trade-offs between generality and task specialization. Massive evaluation suites such as MS~MARCO and the Massive Text Embedding Benchmark (MTEB) demonstrate that no single embedding method uniformly dominates across retrieval, clustering, reranking, and other downstream tasks; instead, model architecture, training objective, and fine-tuning details all matter for the geometry and utility of the resulting embedding space. \citep{nguyen2016msmarco,muennighoff2023mteb}

At the same time, the rise of very large language models (LLMs) as embedding backbones has created a new set of design choices. LLM-based encoders can produce rich, compositional representations, but adapting them to a downstream task is increasingly done with parameter-efficient fine-tuning (PEFT) methods such as LoRA that inject low-rank updates rather than re-training all parameters. While PEFT dramatically reduces computational and storage cost, prior work has also exposed fragility in the fine-tuning process (e.g., sensitivity to initialization, data order, and hyperparameters) and a tendency for task-specific updates to ``overwrite'' aspects of the pretrained geometry that supported broader generalization. \citep{devlin2019bert,hu2022lora,ding2023peft,dodge2020finetune, andreassen2021robustnessfinetuning}

These observations have motivated research on \emph{model merging} and weight-averaging techniques that aim to reconcile the gains of fine-tuning with the robustness of pretrained checkpoints. Approaches range from stochastic weight averaging (SWA) and other trajectory-averaging schemes that seek flatter optima, to \emph{model soups} (simple weight averages of multiple fine-tuned checkpoints) that often improve accuracy and robustness without extra inference cost. More recent surveys and methodological work analyze when and why weight-space interpolation and merging succeed, and how their geometry ties to generalization. \citep{izmailov2018swa,wortsman2022modelsoups,yang2024modelmerging}

Geometric interpolation in weight or latent spaces is a natural further step. Spherical linear interpolation (SLERP) — originally introduced in computer graphics for smooth interpolation on the unit sphere — provides a principled way to interpolate between normalized vectors along great-circle arcs. In model-merging and checkpoint interpolation contexts, SLERP-based merging can preserve directional structure that naive linear averaging might distort; this geometric perspective suggests SLERP as a promising tool for blending pretrained and task-specific weight components while better preserving the base-model manifold. \citep{shoemake1985slerp}

Despite these strands of work, important gaps remain. Past benchmarks and analyses focus largely on semantic textual similarity, retrieval, and classification; far less is known about how different embedding backbones (LLM-based vs.\ non-LLM encoders) represent abstract, non-linguistic structure, or how PEFT-induced specialization interacts with merging strategies that use spherical interpolation. In this paper we aim to fill that gap by studying embedding generalization on a narrow, controlled domain (numerical sequences with provable number-theoretic labels) and by comparing: (i) non-LLM encoders, (ii) LLM backbones adapted with LoRA, (iii) LoRA checkpoints merged via model soups, and (iv) LoRA checkpoints merged via SLERP-based interpolation. Our experiments quantify clustering separability (Silhouette, Davies–Bouldin), inspect KMeans labels for emergent structure, and visualize latent geometry; together they show how SLERP merging recovers base-model structure while retaining much of the task-specific improvement introduced by adapters.

\section{Related Works}

\subsection{Classical and Modern Text Embeddings}
Early work on dense semantic representations focused primarily on static word
embeddings learned through co-occurrence statistics, with GloVe
\cite{pennington2014glove} and word2vec \cite{mikolov2013distributed} establishing
that simple distributional objectives yield surprisingly powerful geometric
representations. Subsequent research extended these ideas to sentence-level
representations, including unsupervised baselines based on weighted bag-of-words
\cite{arora2017simple}.

The introduction of transformer encoders fundamentally shifted the landscape.
Sentence-BERT \cite{reimers2019sentence} demonstrated that Siamese
fine-tuning of pretrained transformers yields embeddings far superior to pooling
raw BERT outputs \cite{devlin2019bert}. Contrastive learning further improved
representation isotropy and generalization, as exemplified by SimCSE
\cite{gao2021simcse}. More recent families of embedding models, such as E5
\cite{wang2024e5} and GTE \cite{li2023generaltextembeddingsmultistage}, scale contrastive objectives while
incorporating multilingual and task-adapted training. Large-scale evaluation
frameworks such as MTEB \cite{muennighoff2023mteb} and MS MARCO
\cite{nguyen2016msmarco} highlight that representational quality varies
significantly across model families, architectures, and fine-tuning strategies.

\subsection{LLM-Based Embeddings and Parameter-Efficient Adaptation}
As large language models became prominent, a growing body of work explored
their use as embedding backbones. While LLMs capture rich hierarchical
structure, adapting them to specialized tasks is typically done through
parameter-efficient fine-tuning (PEFT) methods. Low-Rank Adaptation (LoRA)
\cite{hu2022lora} remains the dominant PEFT technique, introducing structured
low-rank updates that achieve high performance with minimal trainable
parameters. Broader studies on PEFT strategies \cite{ding2023peft} reveal both
their strengths and their fragility, including sensitivity to initialization,
data ordering, and task-specific overfitting. This motivates techniques that
mitigate the dominance of task-specific updates while preserving the underlying
generalization behavior of the base model.

\subsection{Model Merging, Weight Averaging, and Interpolation}
Several lines of research investigate how merging or interpolating model
checkpoints improves robustness and generalization. Stochastic Weight Averaging
(SWA) \cite{izmailov2018swa} demonstrated that averaging weights along the
optimization trajectory leads to flatter minima and improved generalization.
Model soups \cite{wortsman2022modelsoups} extended this concept by averaging
multiple fine-tuned checkpoints, often improving accuracy without modifying
training procedures. More recent surveys (e.g., \cite{yang2024modelmerging})
analyze when merging is effective and how weight-space geometry influences its
success.

Interpolation-based merging introduces a geometric perspective. Linear
interpolation in weight space often performs well, but can distort directional
structure. Spherical linear interpolation (SLERP), originally introduced for
quaternion rotation in computer graphics \cite{shoemake1985slerp}, has recently
been explored as a more principled method for interpolating normalized
parameters or latent vectors. SLERP has gained renewed interest as a merging
strategy because it preserves angular structure, thereby mitigating catastrophic
misalignment that can occur with naive linear averaging.

\subsection{Embedding Structure Beyond Natural Language}
Although embeddings are typically evaluated on natural-language tasks, several
studies explore how models encode abstract or non-linguistic structure. Prior
work demonstrates that transformer embeddings sometimes capture latent
mathematical properties in sequences or symbolic data \cite{saxton2019math}.
However, systematic analyses of number-theoretic structure in embeddings remain
scarce. This gap motivates our investigation: comparing LLM-based and non-LLM
embeddings on controlled numerical-sequence tasks, and evaluating the ability of
SLERP-based merging techniques to recover generalizable structure after
task-specific adaptation.

\section{Model Overview}
\label{sec:model-overview}

\begin{table}[H]
\centering
\scriptsize
\renewcommand{\arraystretch}{1.05} 
\label{tab:architectures-training}
\begin{adjustbox}{max width=\textwidth}
\begin{tabularx}{\textwidth}{
  >{\RaggedRight\arraybackslash}X 
  >{\RaggedRight\arraybackslash}X 
  >{\RaggedRight\arraybackslash}X 
  >{\RaggedRight\arraybackslash}X
}
\toprule
\textbf{Model} & \textbf{Base Architecture} & \textbf{Sentence Embedding Formation} \\
\midrule

\textbf{EmbeddingGemma (0.3B)} \cite{vera2025embeddinggemmapowerfullightweighttext} &
Encoder-only Transformer (Gemma-3 encoder; full bidirectional attention) &
Mean pooling over final encoder hidden states; MRL projection options \\

\addlinespace
\textbf{gte-multilingual-base (0.3B)} \cite{li2023generaltextembeddingsmultistage}&
Encoder-only Transformer (GTE multilingual; XLM-R–style with RoPE + 8k context) &
[CLS] or first-token embedding (projected via MRL) \\

\addlinespace
\textbf{gte-large-en-v1.5 (0.4B)} \cite{li2023generaltextembeddingsmultistage} &
Encoder-only Transformer (GTE English-specialized) &
[CLS] token or projected representation \\

\addlinespace
\textbf{nomic-embed-text-v2-moe (0.5B)}\cite{nussbaum2025trainingsparsemixtureexperts} &
Mixture-of-Experts Transformer (8 experts, 475M total, 305M active) &
Mean pooling; supports multiple output sizes through matryoshka projections\\

\addlinespace
\textbf{Multilingual-e5-large (0.6B)}\cite{wang2024multilinguale5textembeddings} &
Encoder-only Transformer (XLM-R-large backbone; 24 layers) &
Mean pooling of last-layer hidden states  \\

\addlinespace
\textbf{Qwen3 Embedding (0.6B / 4B / 8B)} \cite{qwen3embedding2025}&
Decoder-only Transformer (Qwen3 architecture; causal attention) &
Hidden state of the final \texttt{[EOS]} token after concatenating instruction + query \\

\addlinespace
\textbf{gte-Qwen2-1.5B-instruct} \cite{li2023generaltextembeddingsmultistage} &
Decoder-only Transformer (Qwen2-1.5B backbone) &
Final \texttt{[EOS]} token state \\

\addlinespace
\textbf{GritLM-7B} \cite{muennighoff2025} &
Decoder-only Transformer (Mistral-7B instruction-tuned) &
Likely last-token embedding (not publicly specified) \\

\addlinespace
\textbf{gte-Qwen2-7B-instruct} \cite{li2023generaltextembeddingsmultistage} &
Decoder-only Transformer (Qwen2-7B backbone) &
Final \texttt{[EOS]} embedding \\

\addlinespace
\textbf{e5-mistral-7b-instruct} \cite{wang2023improving} &
Decoder-only Transformer (Mistral-7B, 32 layers, 4096 dim) &
Hidden state of final token after instruction prefix. \\

\addlinespace
\textbf{Llama-Embed-Nemotron (8B)}\cite{nemotron2025} &
Bidirectional Transformer (Llama-3.1-8B with causal mask removed) &
Global average pooling over final-layer hidden states \\

\addlinespace
\textbf{bge-multilingual-gemma2 (9B)} \cite{bge-m3} &
Decoder-only Gemma-2 9B adapted for embeddings &
Last-token embedding with linear projection \\

\bottomrule
\end{tabularx}
\end{adjustbox}
\caption{Architectural differences for all evaluated models.}
\end{table}

\section{Methods}

\subsection{Sequence Construction}
We evaluate embedding models using a controlled numeric–sequence clustering task. Six sequence families were generated: consecutive integers, even numbers, odd numbers, prime numbers, Recamán sequence values, and composite numbers. For each family, we produced $4$ non-overlapping sequences of length $50$, resulting in $24$ total examples. Each sequence was converted into a comma-separated string (e.g., ``2, 4, 6, 8, \dots'') with a maximum length cap of 20k characters to ensure compatibility with all text-based embedding models.

\subsection{Models}
We tested a diverse set of open-source embedding models, including encoder-only models and LLM-based embedding variants (e.g., \texttt{bge-multilingual-gemma2}, \texttt{gte-multilingual-base}, \texttt{Qwen3-Embedding-8B}, \texttt{e5-mistral-7b-instruct}, and others). Models were loaded through the \texttt{SentenceTransformer} interface with \texttt{trust\_remote\_code=True} to ensure each model’s native pooling strategy was used. The environment in which the experiment was performed is Google Colab.

\subsection{Embedding Procedure}
Sequences were encoded in batches of 64 using GPU acceleration. The final embedding matrix $X \in \mathbb{R}^{24 \times d}$ (where $d$ is model-specific) was extracted directly from each model’s recommended embedding output.

\subsection{Dimensionality Reduction}
To visualize global structure before clustering, we projected all embeddings into 2D using t-SNE,using perplexity $=$ 30 (or 10 for fewer than 50 samples) with 1000 iterations. These plots served only as qualitative confirmation of the cluster tendencies.

\subsection{Clustering Evaluation}
Clustering performance was quantified using two standard metrics applied directly to the embedding matrix $X$:

\begin{itemize}
    \item \textbf{Silhouette Score} (higher = better), computed once using the true class labels and once using labels from KMeans.
    \item \textbf{Davies--Bouldin Index} (lower = better), also computed for both true labels and KMeans labels.
\end{itemize}

True labels correspond to the six sequence families. For the unsupervised condition, we ran \texttt{KMeans(n\_clusters = 6)} with 10 restarts.

\section{Evaluation Metrics}
\label{sec:evaluation-metrics}

\subsection{Clustering Metrics}
To quantify separability, we evaluate each embedding configuration using two cluster-quality indices:
\begin{enumerate}
    \item \textbf{Silhouette Score} \cite{ROUSSEEUW198753}:
    \begin{equation}
        s = \frac{b - a}{\max(a,b)}
    \end{equation}
    where $a$ is the mean intra-cluster distance and $b$ is the smallest mean distance to another cluster. Scores range from $-1$ to $1$, with higher values indicating clearer separation. We compute this metric twice:
    \begin{enumerate}
        \item using the \emph{true} sequence groups as labels,
        \item using cluster assignments obtained via \texttt{KMeans}.
    \end{enumerate}

    \item \textbf{Davies–Bouldin Index (DBI)} \cite{4766909}:  
    \begin{equation}
        \mathrm{DBI} = \frac{1}{K}\sum_{i=1}^{K} 
        \max_{j \neq i} 
        \frac{\sigma_i + \sigma_j}{d(\mathbf{c}_i, \mathbf{c}_j)}
    \end{equation}
    where $\sigma_i$ is the intra-cluster dispersion for cluster $i$ and $d(\mathbf{c}_i, \mathbf{c}_j)$ is the distance between cluster centroids. Lower values indicate superior cluster separation.
\end{enumerate}
The combination of silhouette and DBI provides robust insight into whether a model organizes sequences into semantically coherent regions of the embedding space.

\section{Results}

\subsection{Clustering Performance Across Models}

\begin{table}[H]
\centering
\footnotesize
\caption{Clustering results}
\begin{adjustbox}{max width=\textwidth}
\begin{tabular}{l c c c c}
\toprule
{Model Name} & {silhouette true groups} & {davies bouldin true groups} & {silhouette kmeans} & {davies bouldin kmeans} \\
\midrule
EmbeddingGemma(0.3B)& 0.0339 & 2.1082 & 0.0857 & 1.6192 \\
gte multilingual-base(0.3B) & -0.0837 & 3.1835 & 0.1904 & 1.2395 \\
gte-large-en-v1.5(0.4B) & -0.0924 & 3.1173 & 0.1718 & 1.2234 \\
nomic-embed-text-v2-moe(0.5B) & -0.1141 & 3.2009 & 0.2497 & 1.1364 \\
Multilingual-e5-large(0.6B) &  -0.0640 & 2.6140 & 0.1831 & 1.3723\\
Qwen3 Embedding (0.6B)& 0.1830 & 1.5173 & 0.1898 & 1.3560 \\
gte-Qwen2-1.5B-instruct & 0.0563 & 2.0755 & 0.1499 & 1.3922 \\
Qwen3 Embedding(4B)& 0.2724 & 1.2080 & 0.2724 & 1.2081 \\
GritLM-7B & 0.2017 & 1.3827 & 0.2429 & 1.1863 \\
gte-Qwen2-7B instruct & 0.1229 & 1.6347 & 0.1136 & 1.5824 \\
e5-mistral-7b-instruct & 0.3037 & 1.1061 & 0.3037 & 1.1061 \\
Qwen3 Embedding(8B)& \textbf{0.3103} & \textbf{1.0893} & \textbf{0.3103} & \textbf{1.0893} \\
llama-embed-Nemtron(8B) & 0.0953 & 1.7002 & 0.0963 & 1.6583 \\
bge-multilingual-gemma2(9B) & 0.2568 & 1.2246 &  0.2568 & 1.2246 \\
\bottomrule
\end{tabular}
\end{adjustbox}
\label{tab:cluster-results}
\end{table}

Table~\ref{tab:cluster-results} summarizes the clustering quality achieved by each embedding model under two evaluation settings: (1)~clustering with \emph{true number-theoretic group labels}, and (2)~unsupervised clustering using \textsc{KMeans}. Two standard metrics are employed: the Silhouette Coefficient (higher is better) and the Davies--Bouldin Index (lower is better).

Across all models, we observe substantial variation in how well embeddings capture structural properties of numeric sequences. The strongest overall performance is obtained by \textbf{Qwen3 Embedding (8B)}, which achieves the highest Silhouette score ($0.3103$) and lowest Davies--Bouldin score ($1.0893$) under both evaluation regimes. Models such as \texttt{e5-mistral-7b-instruct}, \texttt{Qwen3 Embedding (4B)}, and \texttt{bge-multilingual-gemma2 (9B)} also show consistently strong performance, indicating that larger models with well-structured embedding spaces better preserve number-theoretic relationships.

In contrast, smaller models (0.3B–0.6B range) generally obtain low or even negative Silhouette scores when evaluated with true groups. This suggests that these embedding spaces fail to organize numerical sequences in a way that aligns with the underlying mathematical structure. Nevertheless, these same models show improved performance under \textsc{KMeans}, implying that while global structure is weak, local geometric regularities still exist and can be exploited by unsupervised clustering algorithms.

\subsection{Silhouette and Davies--Bouldin Trends}

\begin{figure}[H]
    \centering
    \includegraphics[width=0.7\linewidth]{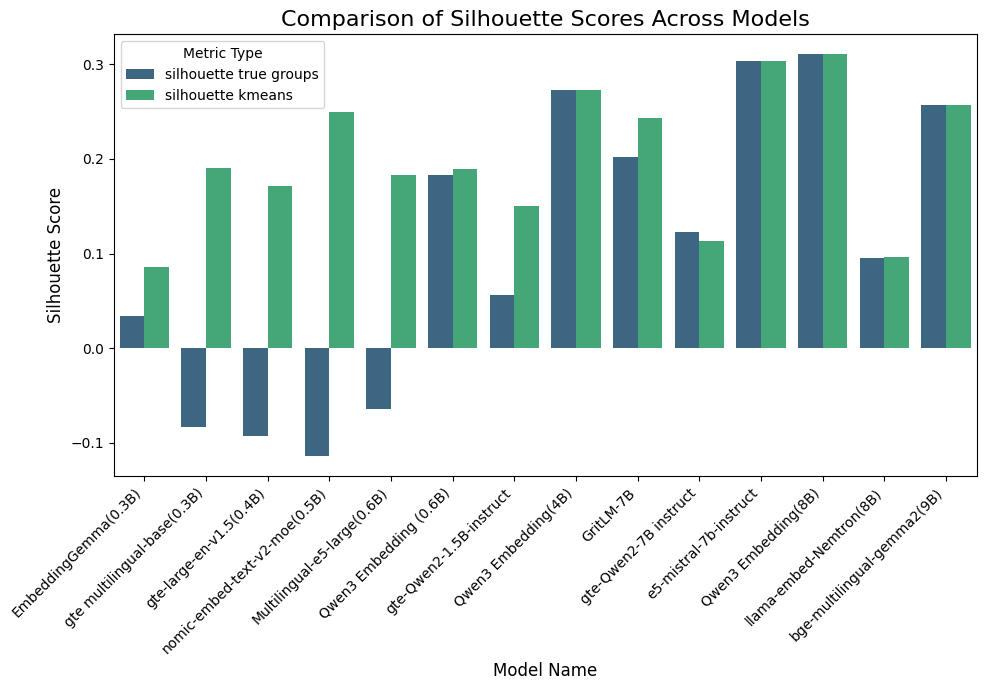}
    \caption{Silhouette scores for all models.}
    \label{fig:silhouette-plot}
\end{figure}

\begin{figure}[H]
    \centering
    \includegraphics[width=0.8\linewidth]{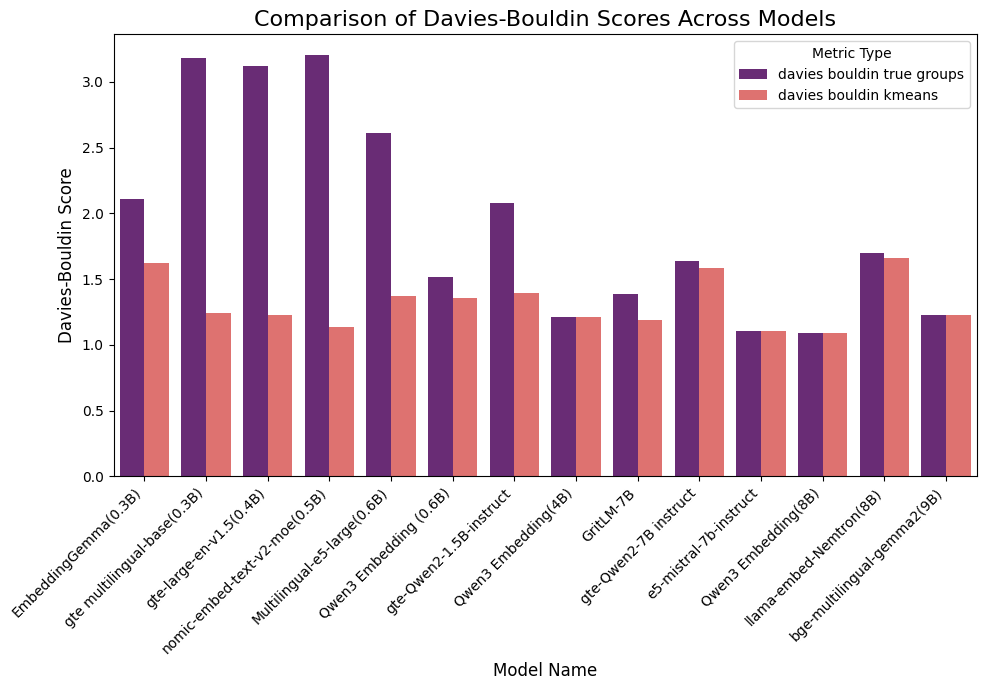}
    \caption{Davies-Bouldin scores for all models.}
    \label{fig:db-plot}
\end{figure}

Figures~\ref{fig:silhouette-plot} and \ref{fig:db-plot} show the Silhouette and Davies--Bouldin metrics for all models. The metrics exhibit a clear monotonic trend: model size and architectural sophistication correlate strongly with clustering quality. Notably, Qwen-based embeddings consistently outperform other models in both metrics, particularly in the larger variants.

The difference between true-label and \textsc{KMeans} scoring further reveals an important behavior:  
\begin{itemize}
    \item When \textsc{KMeans} Silhouette scores exceed those based on true groups, it suggests that the embedding space clusters more naturally according to its own learned geometry rather than according to mathematically defined number-theoretic classes.
    \item This effect is strongest for smaller and mid-sized models, indicating that these models may be forming geometric clusters unrelated to the inherent number-theoretic structure.
\end{itemize}

Such a discrepancy highlights the mismatch between the learned embedding manifold and the intrinsic mathematical properties of the sequences.

\subsection{t-SNE Visualization of Embedding Spaces}

For each model, two-dimensional t-SNE projections were generated to visualize how numeric sequences are arranged in the embedding space, these plots can be found in the appendix section of this paper. Models initialized from LLMs, especially Qwen3 (4B/8B) and e5-mistral-7B, exhibit more distinct and well-separated clusters. These clusters appear smoother, more compact, and more stable.

Models without an LLM base, in contrast, show diffuse and overlapping clusters, confirming the quantitative metrics: their embedding manifolds do not represent number-theoretic properties cleanly. The t-SNE plots also reveal that some models (e.g., \texttt{gte multilingual-base} and \texttt{nomic-embed-text-v2-moe}) produce elongated cluster shapes, suggesting poorly structured latent geometry for this task.

\section{Discussion}

The experimental results highlight several important insights regarding the generalization capabilities of embedding models when tasked with representing number-theoretic structure.

First, the large Qwen3-based embeddings yield the most faithful representations, outperforming all other models across every evaluation metric. This suggests that their training corpus, attention architecture, and embedding design capture hierarchical relationships even in non-linguistic inputs such as sequences of numbers and thanks to the model merging technique they used (SLERP), they managed to keep the base model knowledge even after merging a lot of checkpoints, this sort of generalization shows us just that. This goes to show just how powerful slerp is when it comes to generalization as we can also see some smaller Qwen3 models even out-perform large models which did not use any merging or a different kind of it. On the other hand we also have models that used model souping (EmbeddingGemma and llama-embed-Nemtron) which did not perform as good as the other llm based models, this could be caused by the model souping technique letting LoRA dominate.

Second, models without an LLM-based backbone or those trained on purely text-retrieval objectives struggle in this setting. These models tend to embed numeric sequences in ways that reflect surface-level token or pattern similarities rather than deeper structural properties, resulting in negative Silhouette scores when evaluated against true mathematical groupings.

Third, the contrast between true-group metrics and \textsc{KMeans} metrics highlights that embedding models often impose their own latent structure that may be inconsistent with formal number-theoretic definitions. \textsc{KMeans} sometimes detects clusters that the model naturally prefers—even if these clusters do not correspond to the intended conceptual classes. This reinforces that embedding spaces are not neutral geometric objects but are shaped by the training distributions and inductive biases of the underlying models.

Overall, the study demonstrates that certain LLM-based embeddings, particularly Qwen3 and e5-mistral, exhibit strong potential for representing mathematically meaningful structure in numerical data, while non-LLM embeddings remain limited in this domain. It should be noted though that none of the models have a silhouette score greater than 0.5 which is generally considered to be a good score for this metric but scaling up models like the Qwen3 series seems rather very promising.

\section{Conclusion}

This study examined how a wide range of embedding models
, spanning non-LLM encoders, LLM-based encoders, and their merged variants
, represent structured, non-linguistic inputs in the form of numeric sequences with
well-defined number-theoretic properties. By evaluating models on a controlled,
domain-agnostic clustering task, we isolated representational geometry from
the confounding effects of natural-language semantics. Our findings reveal
consistent and meaningful differences across model families.

LLM-based embedding backbones demonstrate substantially stronger
generalization, producing embedding spaces that better separate numeric
sequence families both in global metrics (Silhouette, Davies--Bouldin) and
in qualitative geometric structure. Larger models, particularly those from
the Qwen3 and e5-mistral families, show the clearest and most stable cluster
boundaries, indicating that the hierarchical and compositional priors learned
during large-scale LLM pretraining extend beyond language and support more
abstract structural reasoning.

Parameter-efficient fine-tuning introduces a trade-off: while LoRA
adapters improve task specialization, they also risk distorting the underlying
geometry responsible for broad generalization. Our results illustrate this
tension directly, with several LoRA-adapted models exhibiting degraded clustering
quality relative to their base models.

Merging strategies play a decisive role in recovering generalization.
SLERP-based interpolation consistently preserves the pretrained directional
structure of the model while integrating task-specific updates, yielding
embeddings that retain both generality and specialization. This contrasts with
naive model averaging and model souping, which in our experiments frequently
allowed adapter components to dominate and erode representational quality.

More broadly, this work demonstrates that embedding models do not merely encode
surface-level token patterns: their latent geometry can meaningfully reflect
the mathematical properties of structured sequences, but only when the underlying
architecture and adaptation strategy support such generalization. The absence
of high Silhouette scores across all models suggests that significant room for
improvement remains, especially in capturing fine-grained number-theoretic
relationships but scaling up embedding models of the same nature as Qwen3 seems promising.

Future work may extend this analysis to larger numerical corpora, symbolic
reasoning tasks, or hybrid architectures that explicitly combine linguistic and
mathematical pretraining. Investigating how SLERP behaves when merging dozens
of checkpoints, or when applied in latent spaces rather than weight space, also
offers promising directions. As embedding applications continue to expand beyond
language, understanding and shaping the geometry of embeddings will become
increasingly critical. This study provides an initial step toward that goal.

\bibliography{references}

\appendix
\section{T-SNE Plots}

\begin{figure}[H]
\centering
\includegraphics[width=0.6\textwidth]{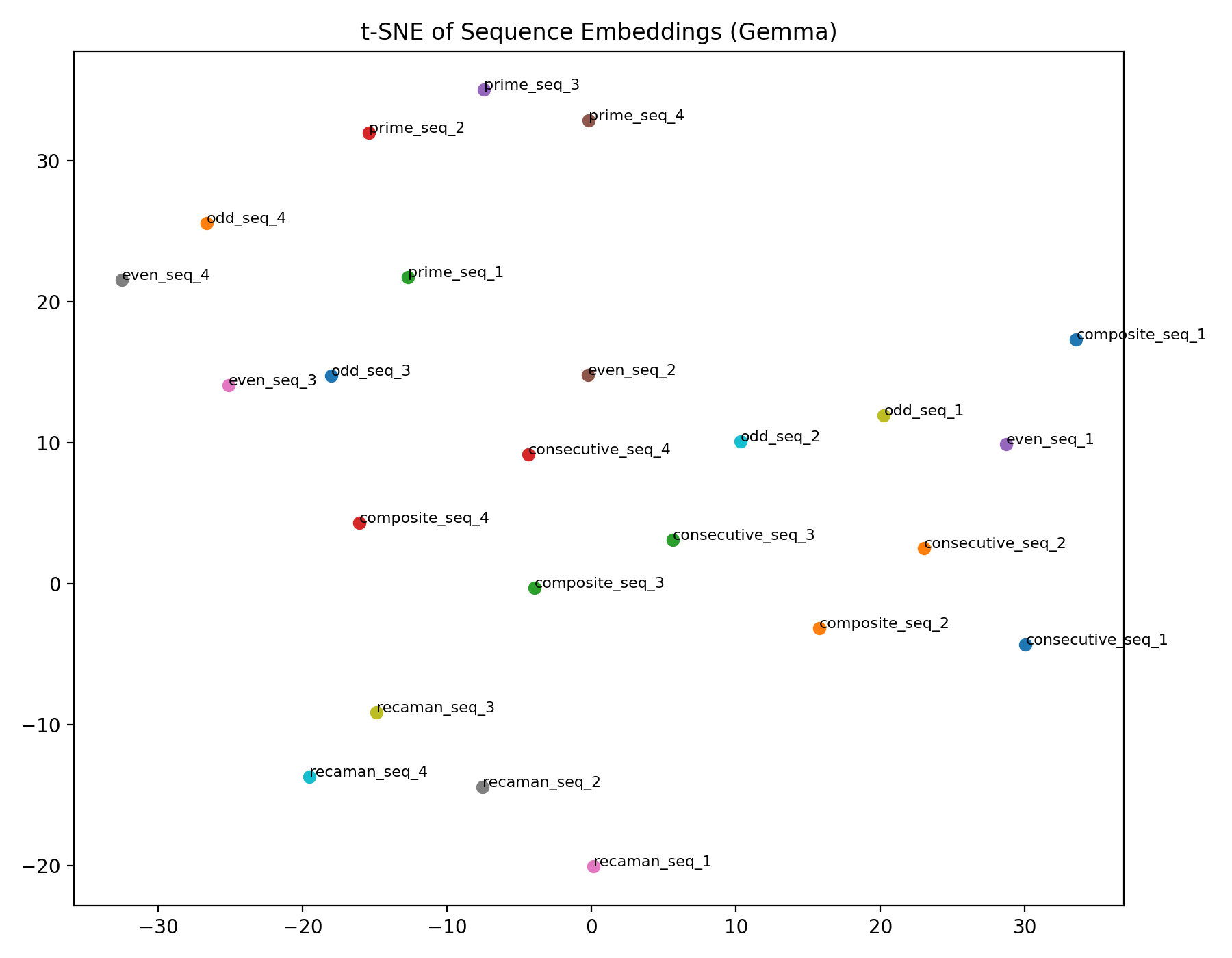}
\caption{EmbeddingGemma Plot}
\end{figure}

\begin{figure}[H]
\centering
\includegraphics[width=0.6\textwidth]{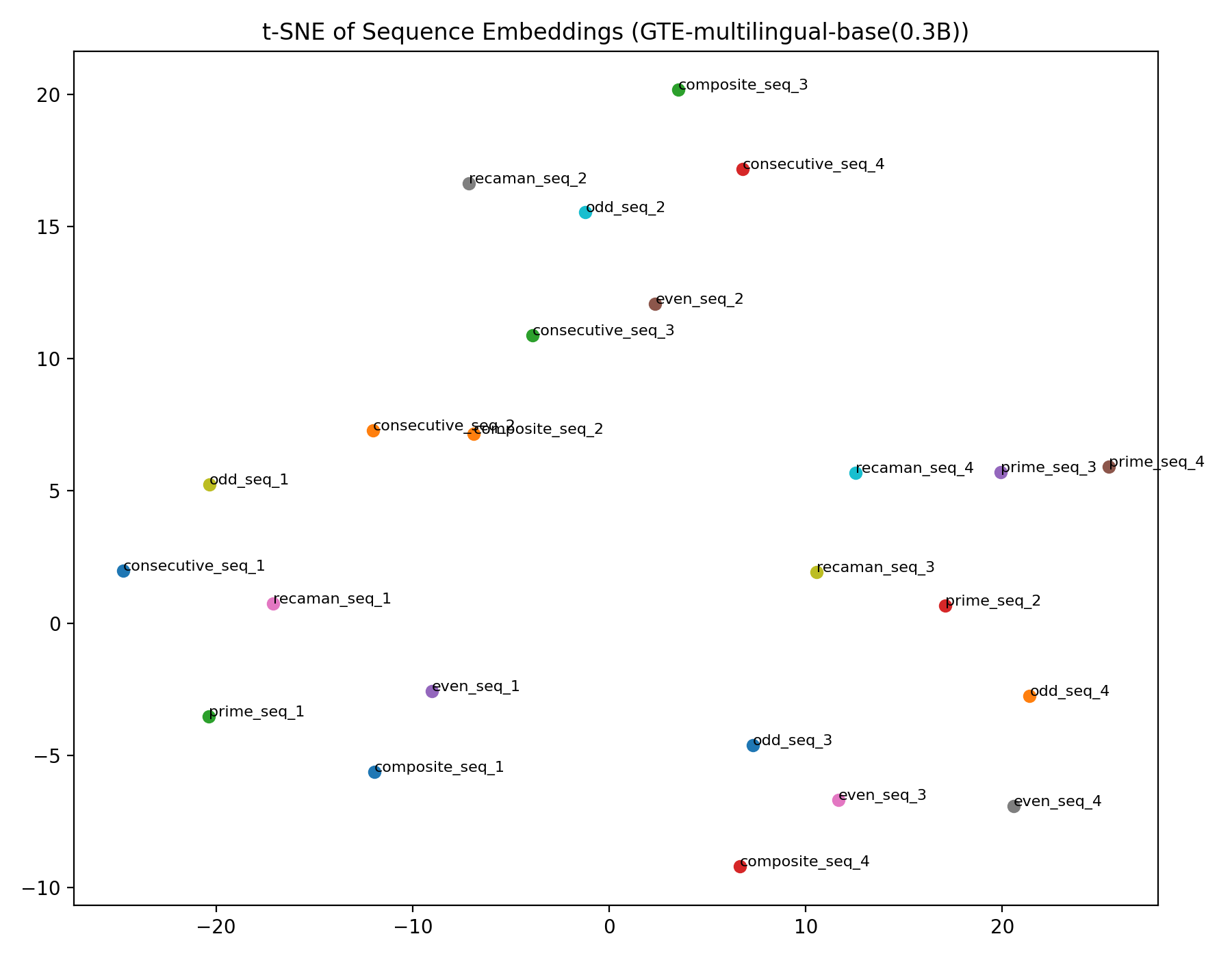}
\caption{GTE Multilingual Base Plot}
\end{figure}

\begin{figure}[H]
\centering
\includegraphics[width=0.6\textwidth]{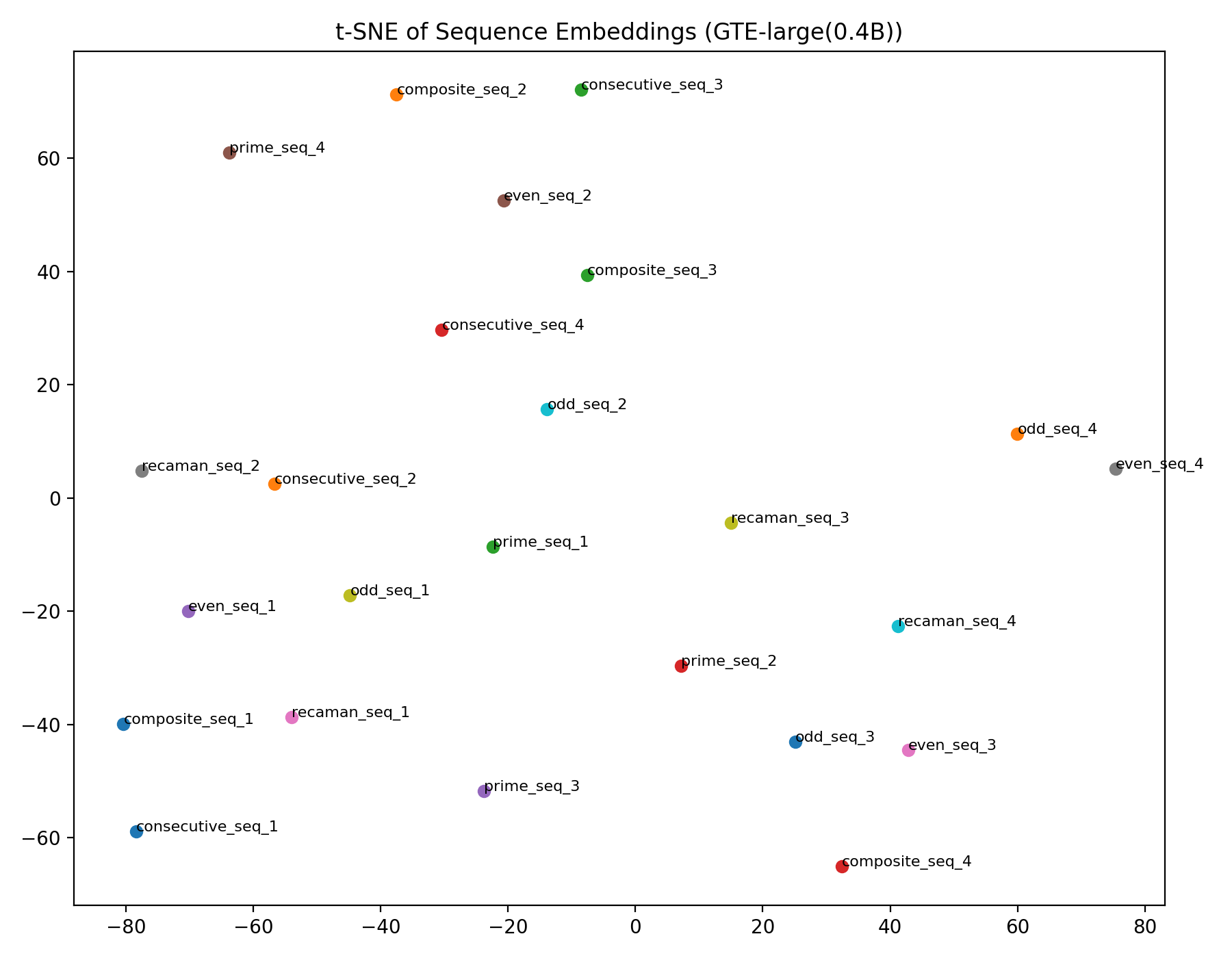}
\caption{GTE-large Plot}
\end{figure}

\begin{figure}[H]
\centering
\includegraphics[width=0.6\textwidth]{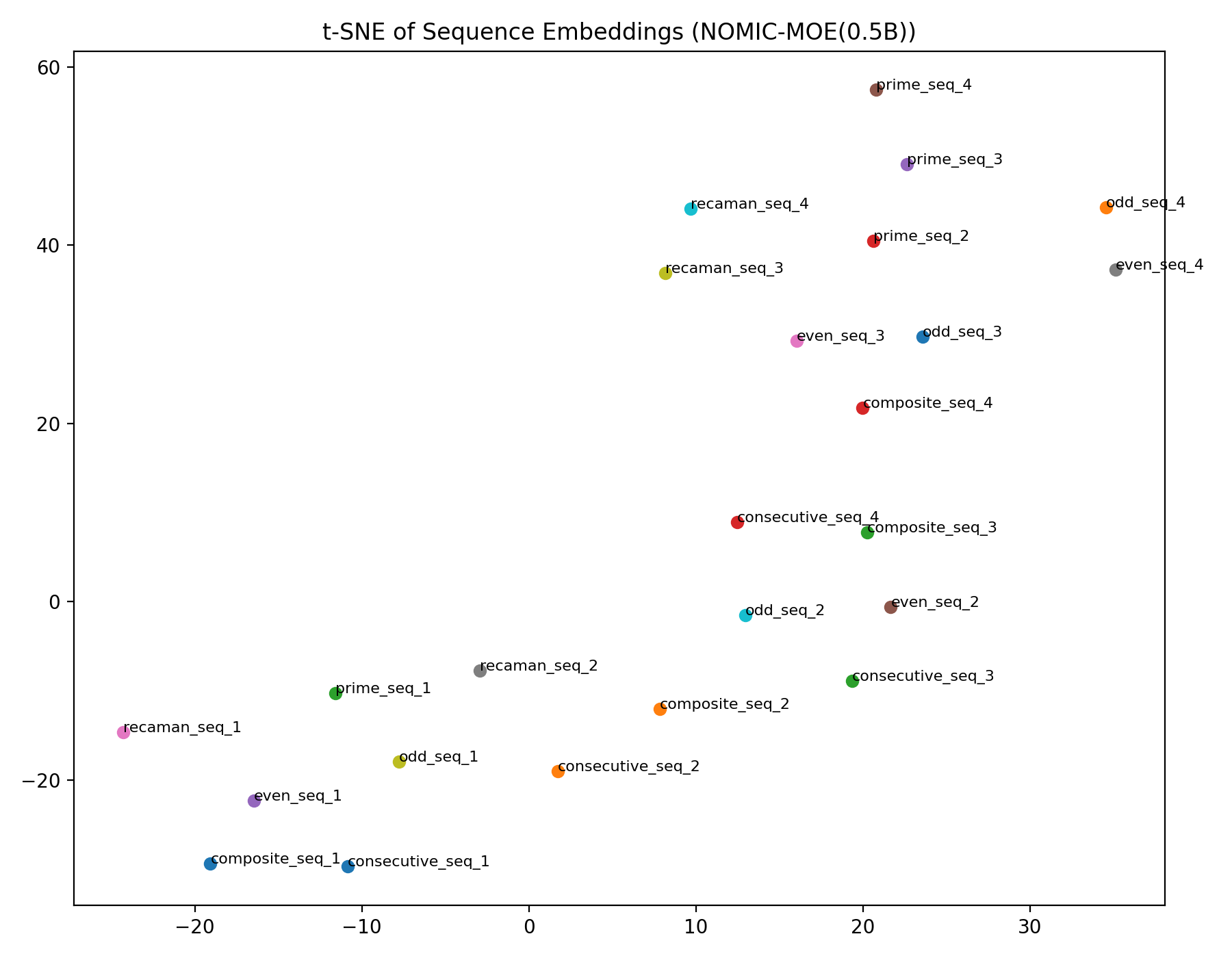}
\caption{Nomic-embed-text-v2-moe Plot}
\end{figure}

\begin{figure}[H]
\centering
\includegraphics[width=0.6\textwidth]{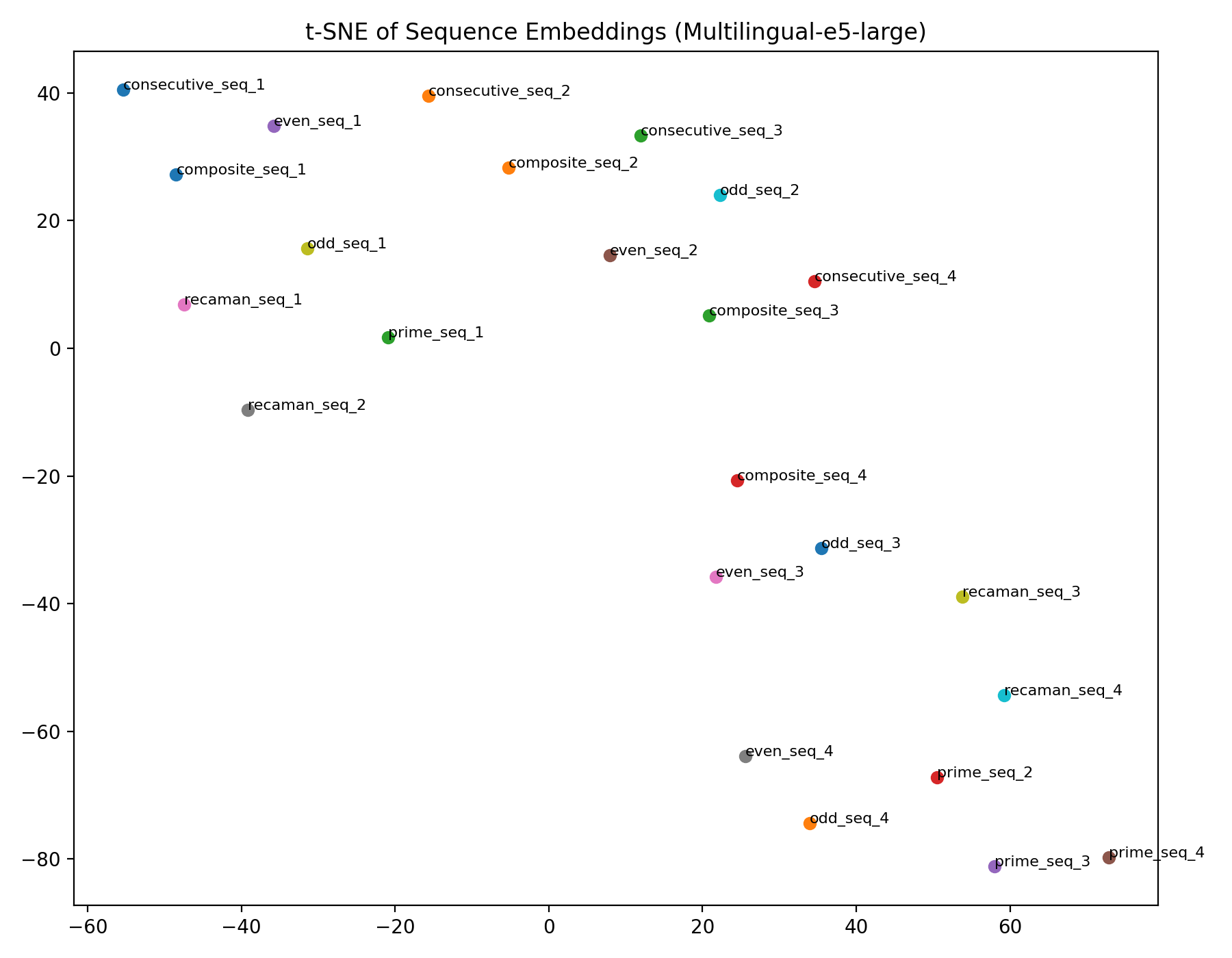}
\caption{Multilingual-e5-large Plot}
\end{figure}

\begin{figure}[H]
\centering
\includegraphics[width=0.6\textwidth]{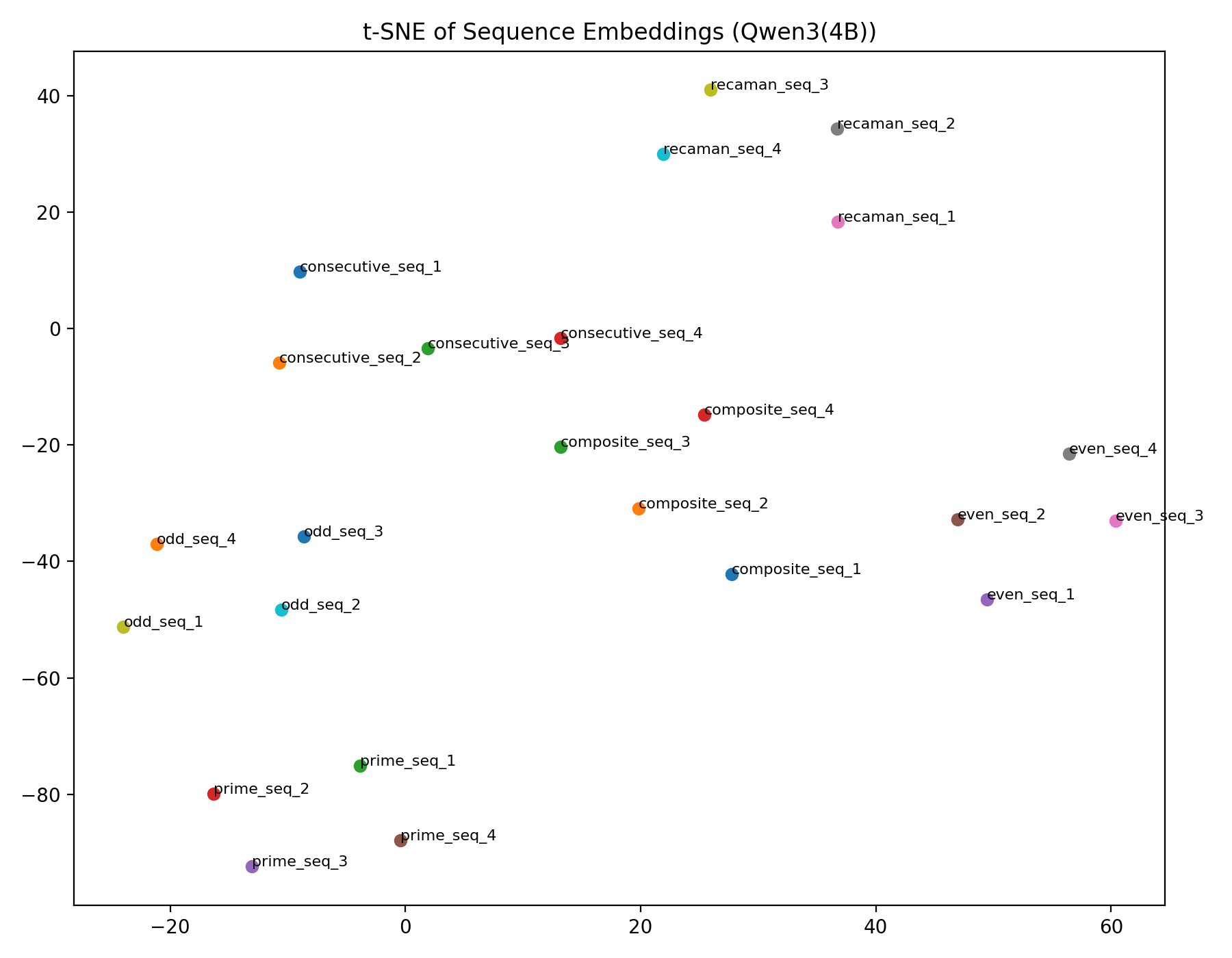}
\caption{Qwen3-Embedding-0.6B Plot}
\end{figure}

\begin{figure}[H]
\centering
\includegraphics[width=0.6\textwidth]{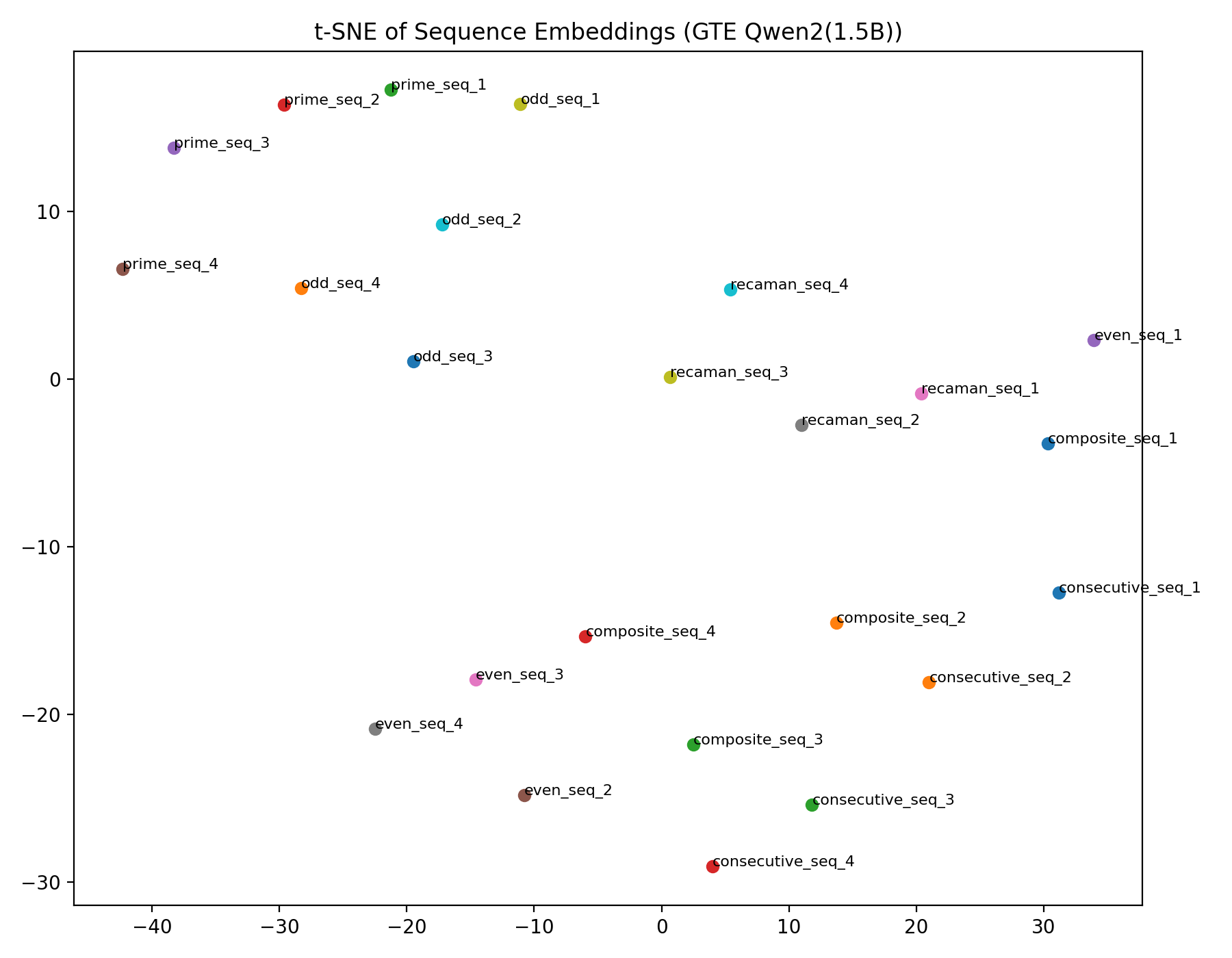}
\caption{GTE-Qwen2-1.5B Plot}
\end{figure}

\begin{figure}[H]
\centering
\includegraphics[width=0.6\textwidth]{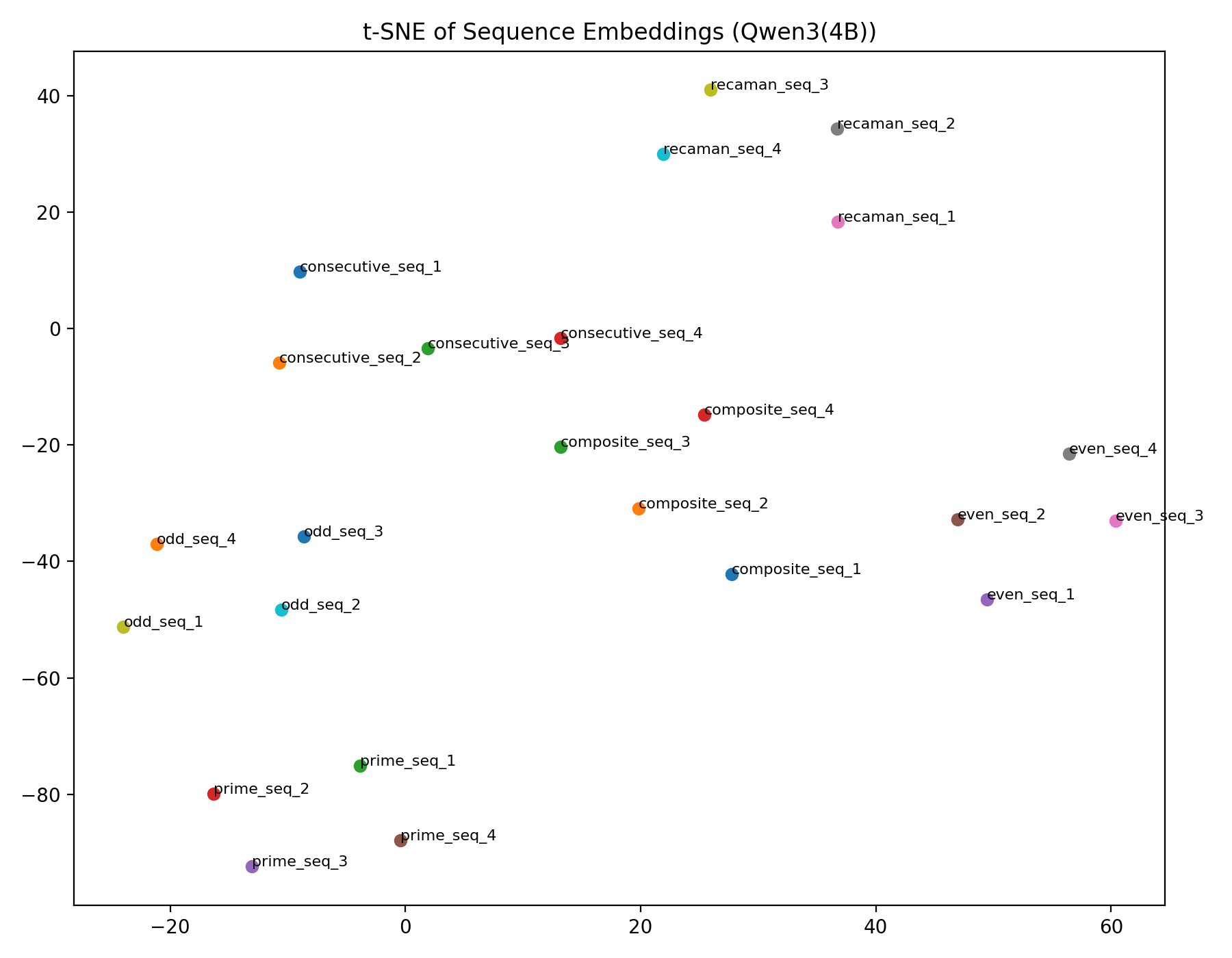}
\caption{Qwen3-Embedding-4B Plot}
\end{figure}

\begin{figure}[H]
\centering
\includegraphics[width=0.6\textwidth]{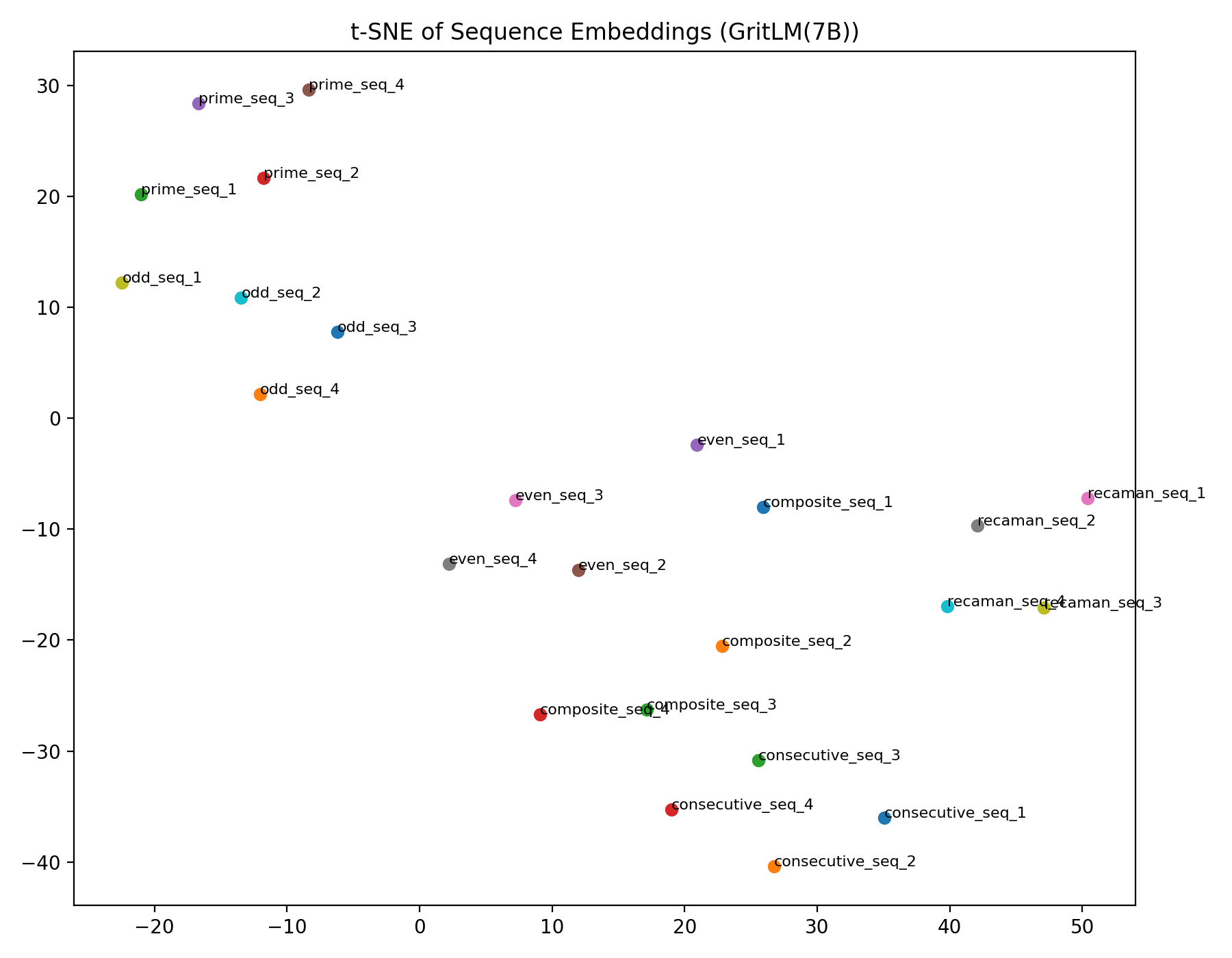}
\caption{GritLM-7B Plot}
\end{figure}

\begin{figure}[H]
\centering
\includegraphics[width=0.6\textwidth]{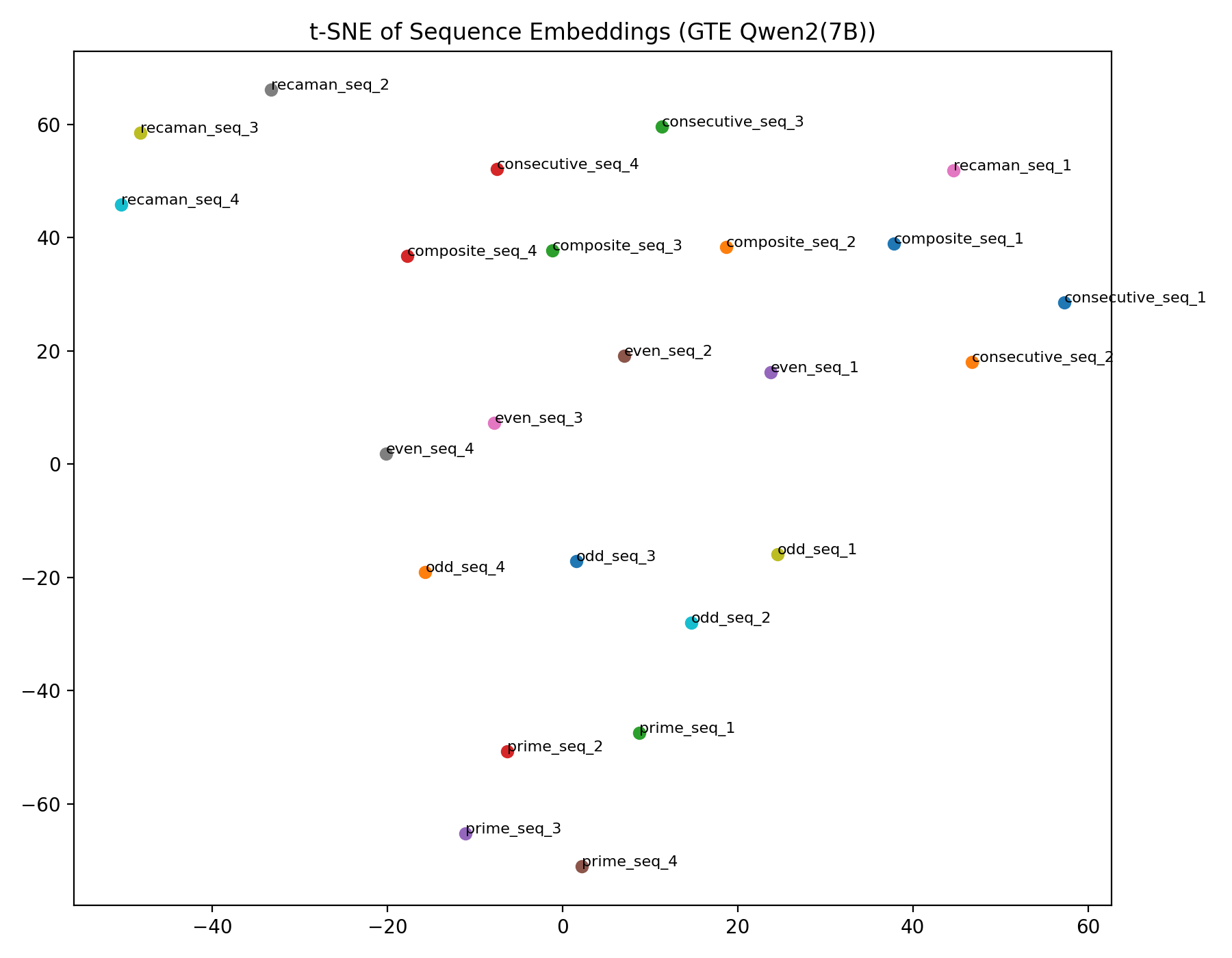}
\caption{GTE-Qwen2-7B Plot}
\end{figure}

\begin{figure}[H]
\centering
\includegraphics[width=0.6\textwidth]{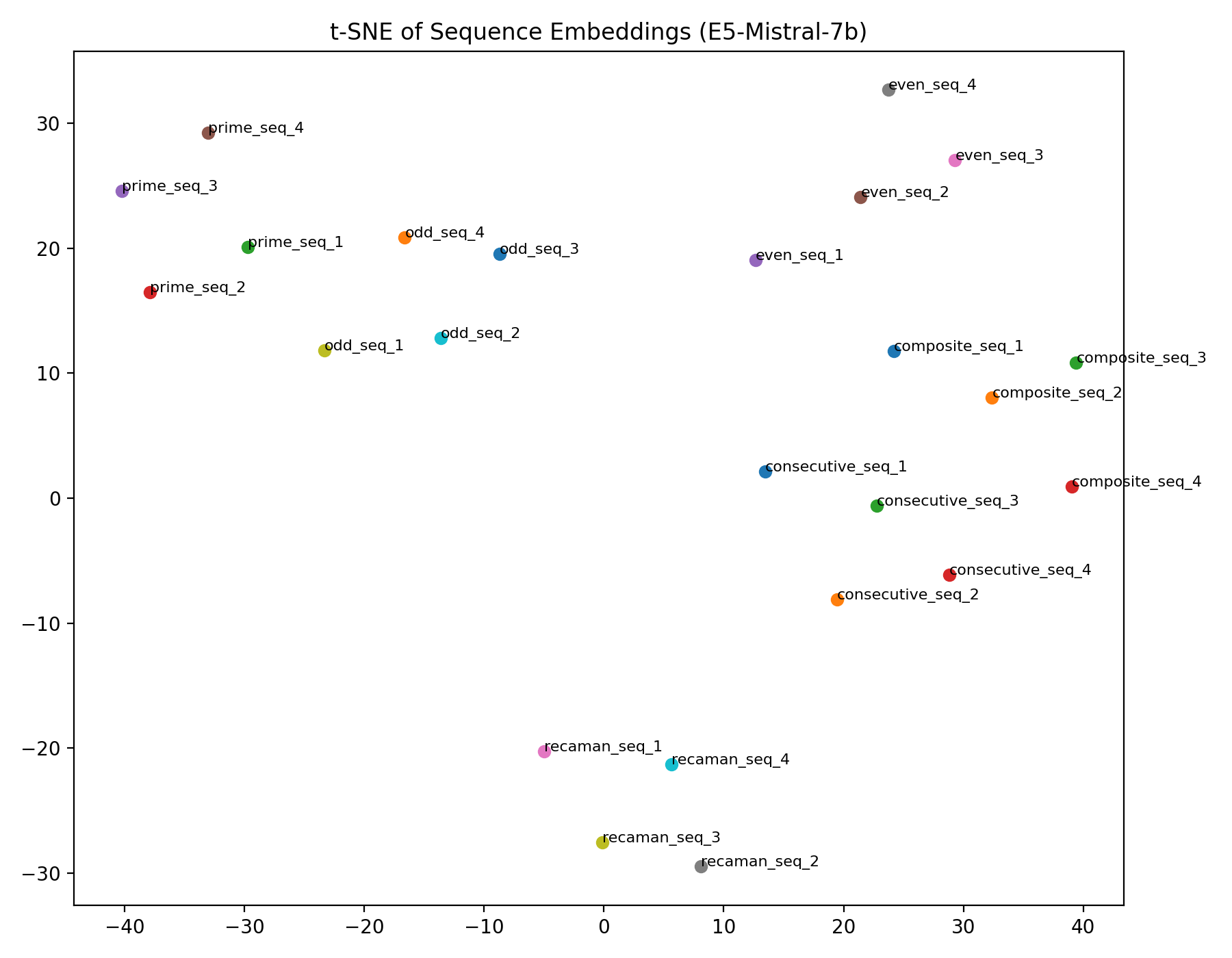}
\caption{E5-Mistral-7b-instruct Plot}
\end{figure}

\begin{figure}[H]
\centering
\includegraphics[width=0.6\textwidth]{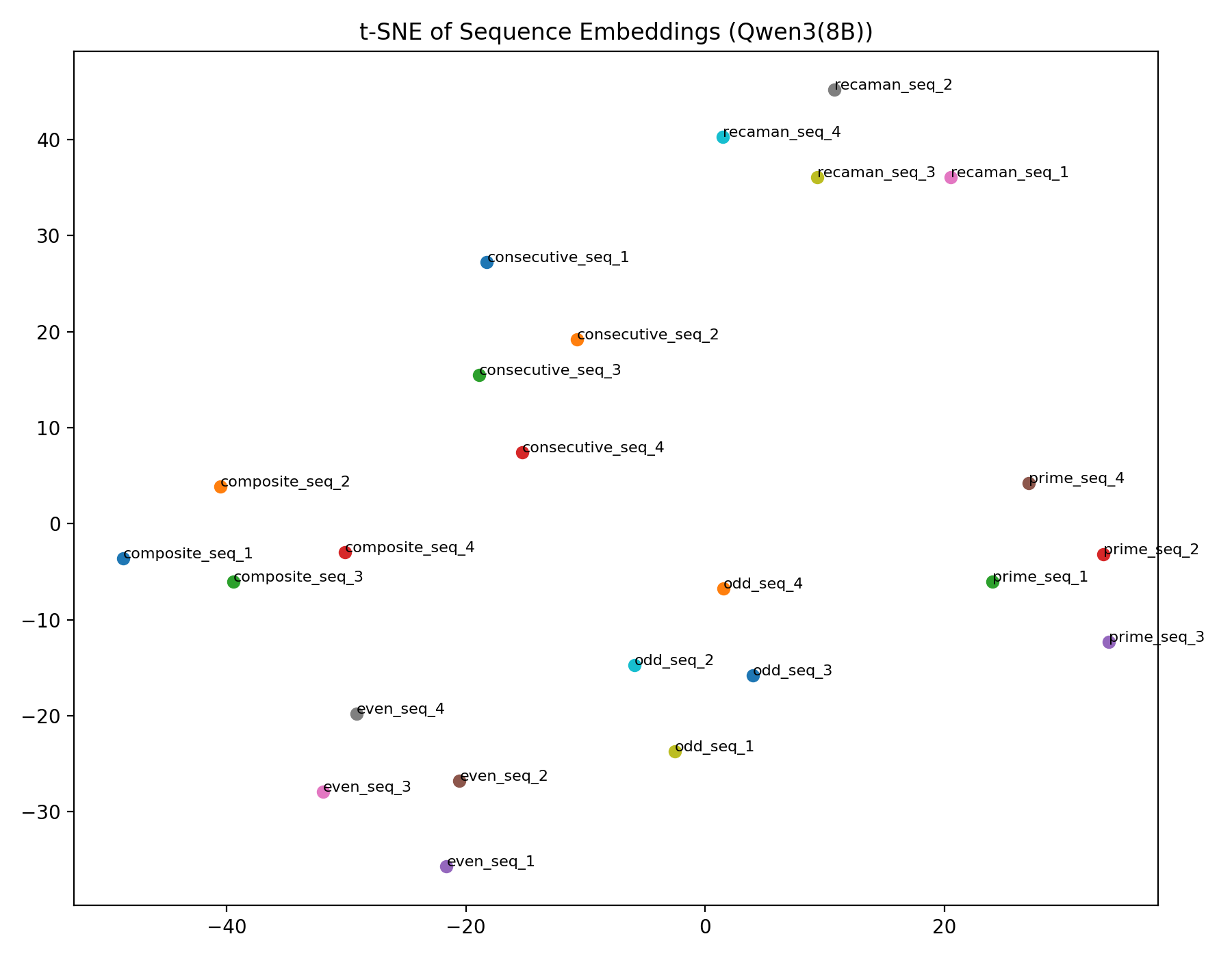}
\caption{Qwen3-Embedding-8B Plot}
\end{figure}

\begin{figure}[H]
\centering
\includegraphics[width=0.6\textwidth]{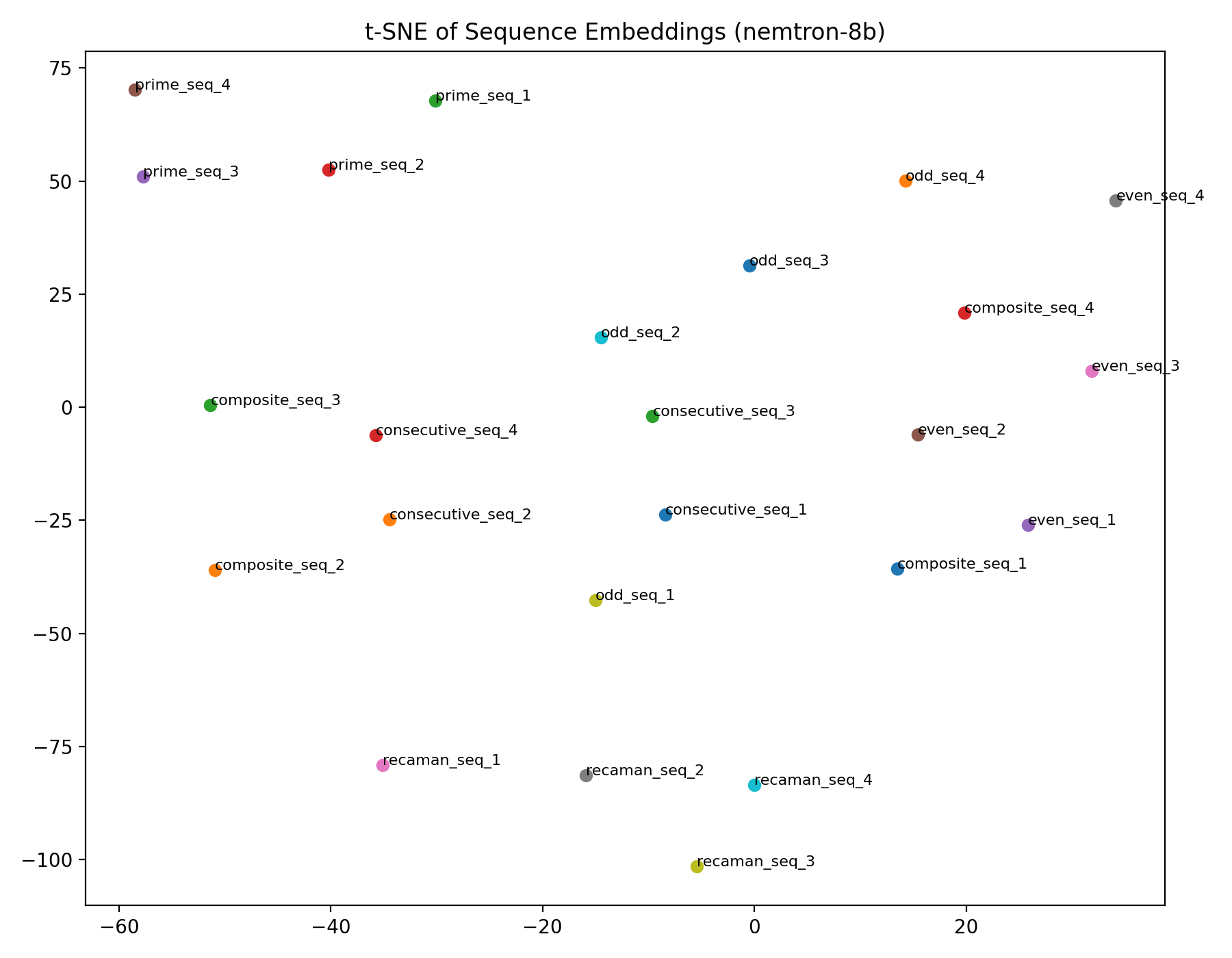}
\caption{llama-embed-Nemtron-8b Plot}
\end{figure}

\begin{figure}[H]
\centering
\includegraphics[width=0.6\textwidth]{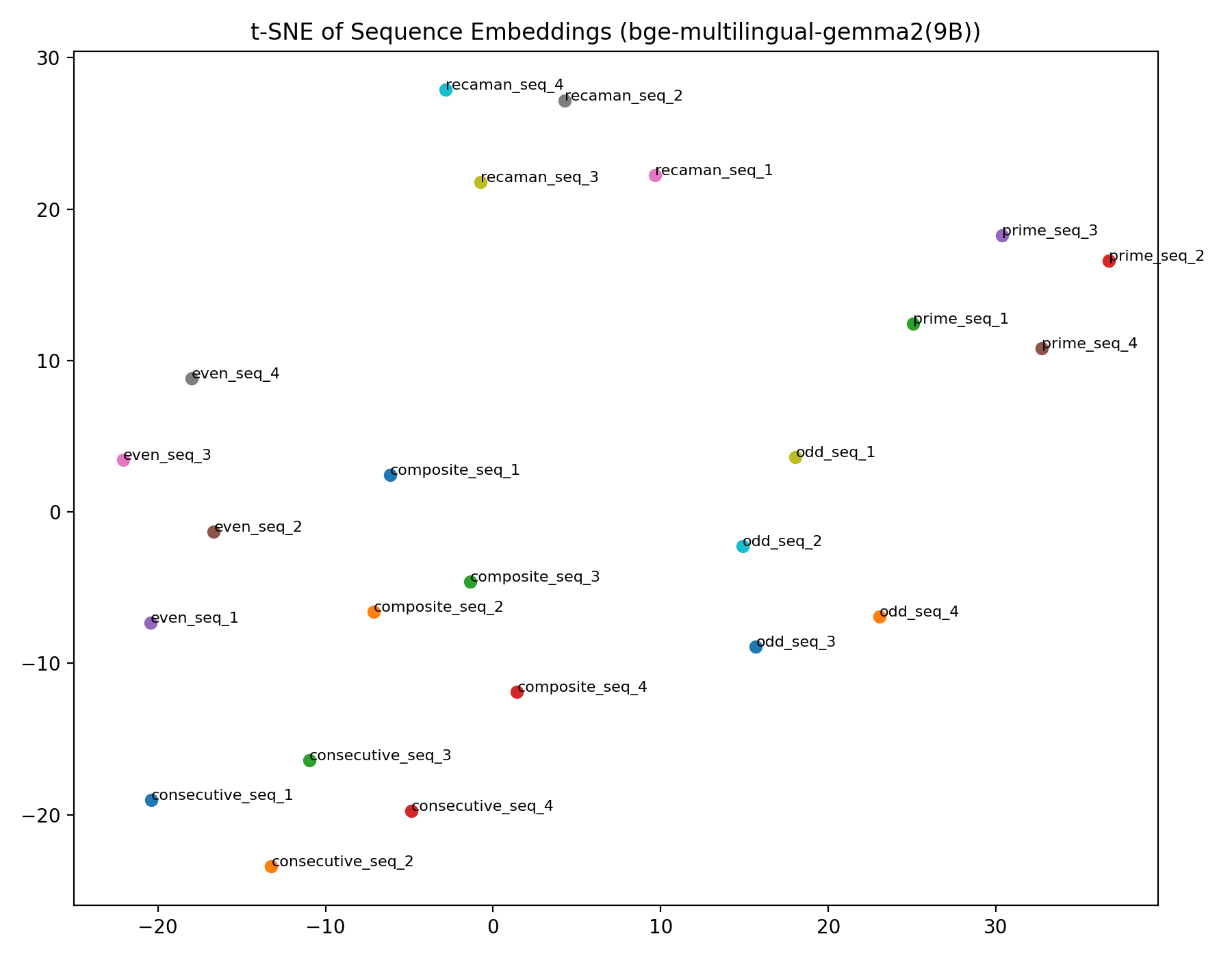}
\caption{bge-multilingual-gemma2(9B) Plot}
\end{figure}

\end{document}